\pdfoutput=1

\documentclass[11pt]{article}

\usepackage{ACL2023}

\usepackage{times}
\usepackage{latexsym}

\usepackage[T1]{fontenc}

\usepackage[utf8]{inputenc}

\usepackage{microtype}

\usepackage{inconsolata}

\usepackage{amsmath}
\usepackage{amssymb}
\usepackage{amsthm}
\usepackage{thmtools, thm-restate}
\usepackage{bm}
\usepackage{tikz}
\usetikzlibrary{calc}
\usepackage{pgfplots}
\usepackage{graphicx}
\usepackage{caption}
\usepackage{subcaption}
\usepackage{float}

\title{Neural Architecture Search for Parameter-Efficient Fine-tuning of Large Pre-trained Language Models}

\author{{\bf Neal Lawton}$^{1}$\footnotemark \quad {\bf Anoop Kumar}$^2$ \quad {\bf Govind Thattai}$^2$ \quad {\bf Aram Galstyan}$^2$ \quad {\bf Greg Ver Steeg}$^2$ \\ $^1$Information Sciences Institute \quad $^2$Amazon Alexa AI \\ \texttt{lawton@isi.edu} \quad \texttt{\{anooamzn,thattg,argalsty,gssteeg\}@amazon.com}}

\begin{document}
\maketitle
\begin{abstract}
Parameter-efficient tuning (PET) methods fit pre-trained language models (PLMs) to downstream tasks by either computing a small compressed update for a subset of model parameters, or appending and fine-tuning a small number of new model parameters to the pre-trained network.
Hand-designed PET architectures from the literature perform well in practice, but have the potential to be improved via automated neural architecture search (NAS).
We propose an efficient NAS method for learning PET architectures via structured and unstructured pruning.
We present experiments on GLUE demonstrating the effectiveness of our algorithm and discuss how PET architectural design choices affect performance in practice.
\end{abstract}

\section{Introduction}
\footnotetext[1]{Work done while at Amazon Alexa AI}

Fine-tuning a large pre-trained language model is a popular method for solving many downstream natural language processing (NLP) tasks. \textit{Full fine-tuning} involves fine-tuning all parameters of the base PLM, resulting in a fine-tuned copy of the model. However, full fine-tuning becomes cumbersome when fine-tuning on multiple downstream tasks due to the massive size of state-of-the-art language models, which range from the millions \cite{devlin2018bert, liu2019roberta} to billions \cite{brown2020language} and now trillions \cite{fedus2022switch} of parameters. Full fine-tuning also carries a risk of \textit{catastrophic forgetting} \cite{jang2021towards, chen2022revisiting}, wherein the PLM's learned useful representation of natural language data is forgotten during fine-tuning.

To address those problems, recent research has focused on  \textit{parameter-efficient tuning} (PET). Rather than fine-tuning all parameters of the base PLM, PET methods choose a small subset of parameters to fine-tune \cite{zaken2021bitfit, guo2020parameter}, or compute compressed parameter updates \cite{hu2021lora, mahabadi2021compacter}, or append and fine-tune a small subset of new parameters \cite{houlsby2019parameter, li2021prefix, hambardzumyan2021warp, he2021towards}. Each of these methods has their own advantages and disadvantages, but one question relevant to all these methods is \textit{which parts of the network are most efficient to fine-tune, and what is the most parameter-efficient way to fine-tune them}? 

Here we answer this question by designing and applying a fine-grain NAS method for learning PET architectures. Our method uses a first order approximation of the loss function and is computationally efficient. We compare our approach with several hand-designed PET methods and find that the architectures learned by our method generally achieve comparable or higher development set performance on GLUE tasks \cite{wang2018glue} for the same number of parameters. We conclude by examining the PET architectures learned by our method and discuss the affect of architecture design choices on parameter efficiency.

\section{Related work}

Many different PET methods exist in the literature. \textit{Adapter networks} insert and fine-tune small adapter modules to a base PLM. \citet{rebuffi2017learning} introduced adapter networks to the visual domain, and \citet{houlsby2019parameter} introduced adapters to transformers. Adapters have been applied to text generation \cite{lin2020exploring}, translation \cite{bapna2019simple}, and multi-task learning \cite{pfeiffer2020mad, pfeiffer2020adapterfusion}. \citet{peters2019tune} compare adaptation with full fine-tuning. AdapterHub \cite{pfeiffer2020adapterhub} enables easy sharing of adapter models. Additionally, \citet{mosbach2020stability} propose best practices for producing strong full fine-tuning baselines.

Prompt-tuning methods fine-tune a PLM by inserting prompt tokens into the input sequence. Continuous prompts \cite{li2021prefix, lester2021power, hambardzumyan2021warp} or discrete prompts \cite{shin2020autoprompt} can be learned or engineered \cite{brown2020language}. \citet{gu2021ppt} demonstrate the effectiveness of pre-training prompts for low resource tasks. 

Some methods fine-tune a subset of parameters \cite{zaken2021bitfit, guo2020parameter}, or compute compressed parameter updates \cite{hu2021lora, mahabadi2021compacter}. These methods fine-tune the PLM without increasing test-time inference latency. \citet{he2021towards} and \citet{mao2021unipelt} combine multiple PET methods.

Beyond parameter-efficient tuning, NAS has previously been used to discover more parameter-efficient base language models. \citet{cheong2019transformers} use magnitude pruning to reduce the number of parameters in BERT. Many efforts at pruning BERT have focused on pruning attention heads from the multi-head attention (MHA) modules \cite{michel2019sixteen, voita2019analyzing, li2021differentiable}. \citet{sajjad2020poor} evaluate different ad-hoc strategies for shrinking the depth of a BERT encoder. \citet{so2019evolved} use an evolutionary NAS method to learn an improved transformer cell. In contrast to NAS, distillation can be used to compress language models \cite{sanh2019distilbert, jiao2019tinybert, sun2020mobilebert}.

In our experiments section, we examine the architectures learned by our algorithm and consider what they say about which parts of the network are most parameter-efficient to fine-tune. \citet{merchant2020happens} explore a similar question, probing the network activations to understand how the network's representation of natural language data changes during full fine-tuning.

\section{Method}
\begin{table*}[!t]
\centering
\begin{tabular}{|c|c|c|c|c|c|c|c|c|c|c|c|}
\hline
Method & \#params & MNLI & SST-2 & MRPC & CoLA & QNLI & QQP & RTE & STS-B & Avg. \\
\hline
FFT & 355M & $90.6$ & $96.0$ & $89.2$ & $66.8$ & $94.6$ & $91.6$ & $85.2$ & $91.5$ & $88.2$ \\
BitFit & 273k & $89.2$ & $95.6$ & $88.2$ & $65.0$ & $93.9$ & $88.1$ & $81.9$ & $91.4$ & $86.7$ \\
Adapters$^\dagger$ & 3.0M & $90.2$ & $96.1$ & $90.2$ & $68.3$ & $94.8$ & $91.9$ & $83.8$ & $92.1$ & $88.4$ \\
\hline
LoRA & 3.4M & $90.7$ & $95.3$ & $89.7$ & $65.1$ & $93.8$ & $90.3$ & $84.8$ & $91.7$ & $87.7$ \\
MaM & 3.4M & $\mathbf{90.6}$ & $95.3$ & $89.7$ & $65.1$ & $93.8$ & $90.3$ & $84.8$ & $91.7$ & $87.7$ \\
S-MaM & 3.4M  & $\mathbf{90.6}$ & $\mathbf{95.9}$ & $90.4$ & $66.3$ & $\mathbf{94.5}$ & $90.6$ & $85.2$ & $91.6$ & $88.1$
 \\
U-MaM & 3.4M & $90.3$ & $95.8$ & $\mathbf{90.7}$ & $\mathbf{66.8}$ & $94.1$ & $\mathbf{90.8}$ & $\mathbf{85.9}$ & $\mathbf{91.8}$ & $\mathbf{88.3}$ \\
\hline
WARP$^\dagger$ & 25k & $88.2$ & $\mathbf{96.0}$ & $\mathbf{90.8}$ & $60.6$ & $\mathbf{93.5}$ & $84.5$ & $\mathbf{75.8}$ & $88.6$ & $\mathbf{84.8}$ \\
S-BitFit & 25k & $84.1$ & $94.2$ & $70.6$ & $40.2$ & $88.9$ & $83.8$ & $56.0$ & $76.8$ & $74.3$ \\
U-BitFit & 25k & $\mathbf{88.8}$ & $95.5$ & $85.3$ & $\mathbf{62.1}$ & $\mathbf{93.5}$ & $\mathbf{87.7}$ & $74.0$ & $\mathbf{90.3}$ & $84.6$ \\
\hline
\end{tabular}
\caption{GLUE development set score for learned and hand-crafted PET architectures. We report the result for WARP$^\dagger$ from \citet{hambardzumyan2021warp} and for Adapters$^\dagger$ from \citet{hu2021lora}.}
\label{table:results_table}
\end{table*}

The architecture search space we choose for our NAS method is based on BitFit \cite{zaken2021bitfit} and LoRA \cite{hu2021lora}, two of the most popular methods for parameter-efficient fine-tuning in the literature. We consider both structured and unstructured variants of each of these, where the non-zero pattern of the learned PET parameters is restricted or unrestricted, respectively. Specifically, our search space consists of the following:

\begin{enumerate}
    \item Learning an update $\Delta b$ for each vector of bias parameters $b$. In \textit{structured bias-tuning}, for each PLM module, the NAS algorithm must choose whether $\Delta b = 0$ or not. In \textit{unstructured bias-tuning}, for each PLM module, the NAS algorithm must choose which components of $\Delta b$ should be zero or non-zero.
    \item Learning a low-rank (LoRA \citealp{hu2021lora}) update $\Delta W = UV^\top$ for each user-specified parameter matrix $W$. The maximum possible rank for the update is also user-specified. In \textit{structured LoRA}, for each parameter matrix $W$, the NAS algorithm must decide what the rank of the update $UV^\top$ should be. In \textit{unstructured LoRA}, the NAS algorithm must decide which components of $U$ and $V$ should be non-zero. 
\end{enumerate}

The collection of updates $\Delta b$ and $\Delta W$ are the PET parameters. In this search space, any number of the above PET modules can be applied to a base PLM without increasing the latency of inference, just like BitFit \cite{zaken2021bitfit} and LoRA \cite{hu2021lora}.

\subsection{Pruning}

We perform NAS via pruning. Our NAS method begins by training a PET architecture of a maximum user-specified size: for each bias tuning module, we fine-tune all bias parameters, and for each LoRA update module, we learn a dense low-rank update with a user-specified rank (in all our experiments, we use rank-16 initial LoRA updates). After training the initial PET architecture, our method decides which PET parameters to prune and which to keep. Then we re-initialize and re-train the pruned architecture before evaluating on the validation set.

The criteria that we use to decide which PET parameters to prune is based on a first-order approximation of the change in training loss that results from pruning a PET parameter $\theta$:
\[
- \theta \cdot \frac{\partial \mathcal L}{\partial \theta}.
\]
Note that this is a common pruning criterion, e.g., see \citet{molchanov2016pruning}. This criterion is straight forward to use when deciding whether to prune a single PET parameter, as in unstructured bias-tuning and unstructured LoRA. For structured bias-tuning, we sum this criterion over the entire bias update $\Delta b$, and for structured LoRA, when considering what column of $U$ and $V$ to prune, we sum the criterion over each column of $U$.

Pruning via evaluating the criterion at the end of training does not yield better-than-random architectures. We observe that the value of the pruning criterion may change drastically from one stochastic gradient descent (SGD) step to the next. To maximally smooth the noise introduced by SGD, we instead average the pruning criterion over all training SGD steps. This yields the most consistent indication of which PET parameters are efficient to prune.

Our NAS algorithm takes as input a parameter budget specifying the desired maximum number of parameters in the learned PET architecture. After training the initial PET architecture and evaluating each pruning criterion, we apply each pruning operation in increasing criterion order until the number of parameters in the PET architecture falls below the parameter budget. This way, pruning operations that are estimated to increase the training loss the least are applied first.

\subsection{Initialization}

Correct initialization is important for successfully applying this algorithm. After pruning, we re-initialize and re-train the learned PET architecture before evaluating on the validation set. We find that it is important to use the same initialization after pruning as before. We believe this is a consequence of the lottery ticket hypothesis \cite{frankle2018lottery}.

We always initialize bias parameter updates as zero, as do other works, and find this works well. However, we find that the initialization for LoRA given in the original paper \cite{hu2021lora}, which initializes the matrix $U$ with zeros and $V$ with a Gaussian distribution, is not ammenable to unstructured LoRA pruning. Because the parameters in the matrix $U$ are initialized zero, the magnitudes of those parameters are likely to remain small throughout training relative to the magnitudes of the parameters in $V^\top$. Consequently, the pruning criterion for unstructured LoRA updates is likely to favor pruning parameters from $U$ over $V$, leading to an unbalanced, parameter-inefficient LoRA update. Instead, following the same reasoning given for Kaiming initialization \cite{he2015delving}, we recommend the following initialization:

\begin{align}
U \sim \mathcal N(0, 1/\sqrt{m}) & & & V \sim \mathcal N(0, 1/\sqrt{n}),
\end{align}

where $m$ is the first dimension of the matrix $U$ (i.e., the "fan-in"), and $n$ is the second dimension of the matrix $V^\top$ (i.e., the "fan-out"). With this initialization, the expected square gradients for the parameters of $U$ and $V$ are equal.

\section{Experiments}
\begin{figure*}[t]
    \centering
    \begin{subfigure}[b]{0.49\textwidth}
        \centering
        \includegraphics[width=\textwidth]{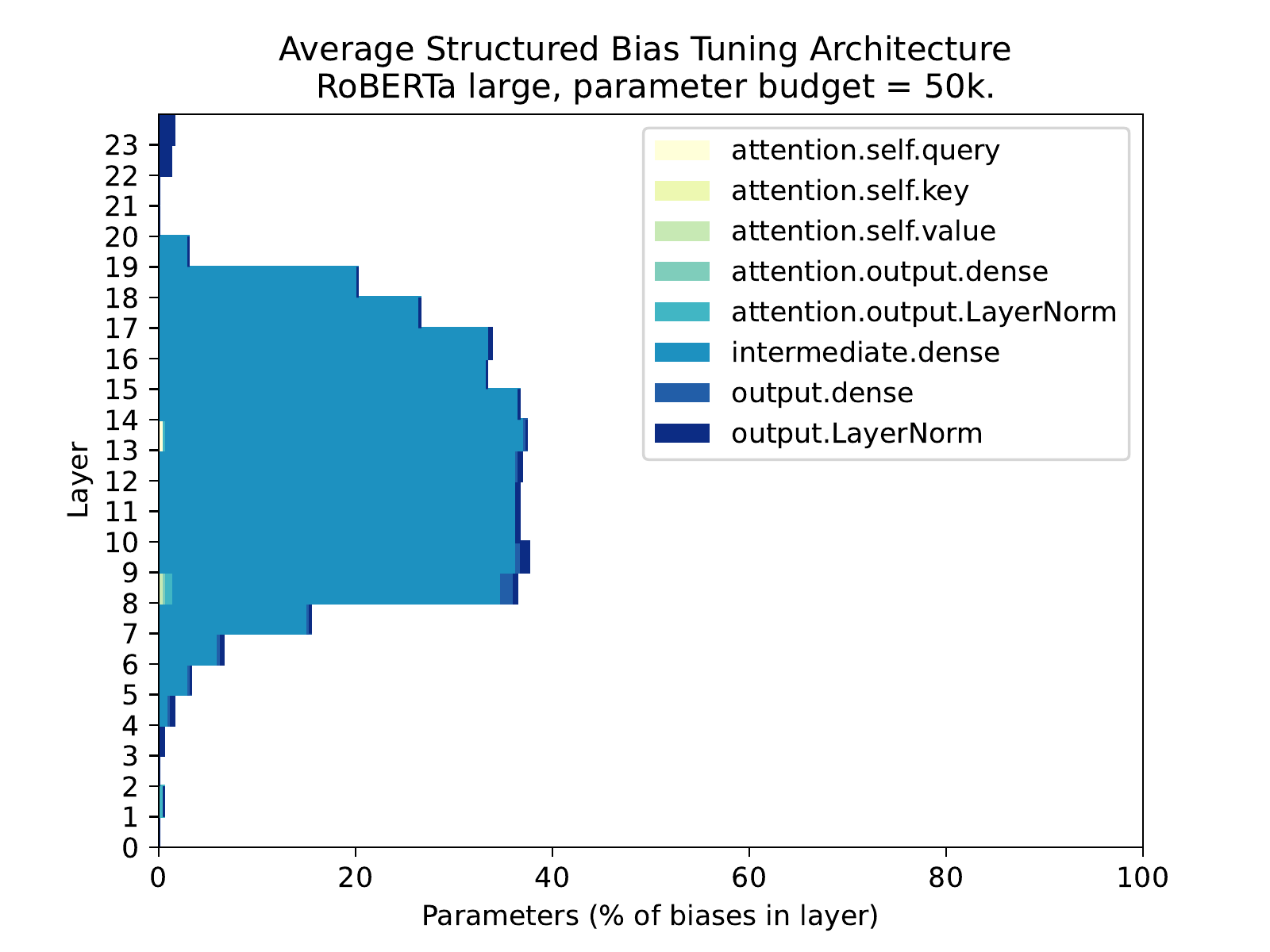}
        \caption{}
        \label{fig:sbt_architecture}
    \end{subfigure}
    \hfill
    \begin{subfigure}[b]{0.49\textwidth}
        \centering
        \includegraphics[width=\textwidth]{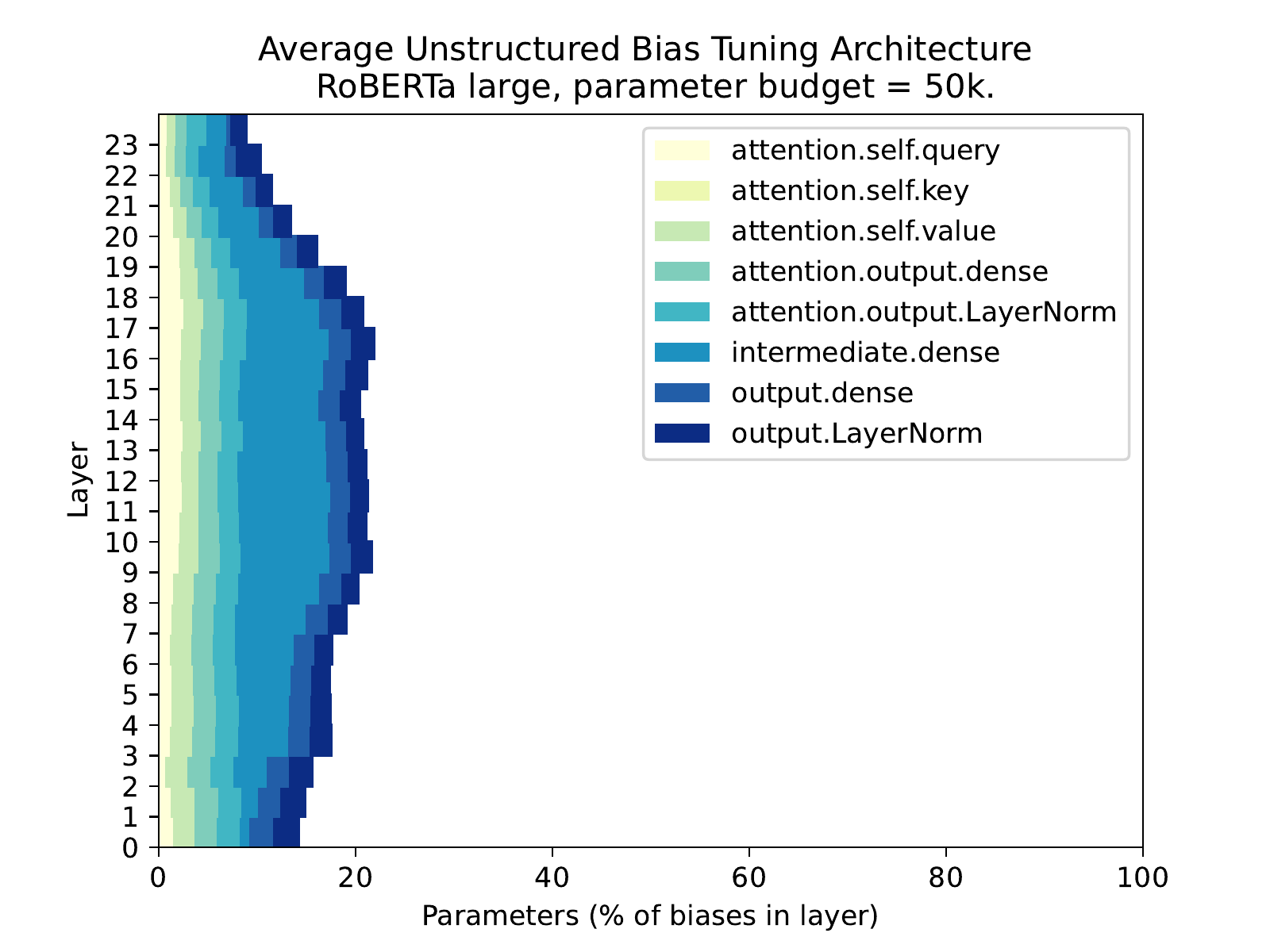}
        \caption{}
        \label{fig:ubt_architecture}
    \end{subfigure}
    \caption{Average learned architecture for (a) structured bias-tuning and (b) unstructured bias-tuning.}
\end{figure*}

Details of our experimental setup, including hyperparameter choices, are available in the appendix. In all experiments we report median validation score at the end of training over 5 random initializations using the GLUE development set for validation.

\subsection{Comparing to Full Fine-tuning}

Here we present results for training larger PET architectures with the aim of achieving performance similar to full fine-tuning, but with fewer parameters. In addition to structured or unstructured bias-tuning, our learned PET architectures add structured or unstructured LoRA updates to the MHA query modules, key modules, and the \verb|dense| feed forward network (FFN) modules. In Table \ref{table:results_table}, our learned structured PET architecture is labeled \verb|S-MaM|, and our learned unstructured PET architecture is labeled \verb|U-MaM|. We compare our method with a LoRA baseline \cite{hu2021lora} and a baseline similar to  Mix-and-Match (MaM) \cite{he2021towards}. Our LoRA baseline fine-tunes all bias parameters and adds rank-$8$ updates to all MHA query and key modules. Our MaM-like baseline fine-tunes all bias parameters and adds rank-$8$ updates to all MHA query and key modules and all FFN modules. 

Results for this experiment with parameter budget 3.4M are in Table \ref{table:results_table}. In our \verb|S-MaM| and \verb|U-MaM| experiments, we prune from an initial architecture with 6.8M parameters. We observe that our \verb|S-MaM| architecture achieves slightly higher average GLUE \cite{wang2018glue} validation score over our MaM-like baseline, and our \verb|U-MaM| architecture slightly higher average GLUE validation score over our \verb|S-MaM| architecture. We conclude that structured architecture search provides a small positive benefit over the uniform-rank baseline architecture, and that unstructured architecture search provides a small positive benefit over structured architecture search. We also observe our \verb|U-MaM| architecture achieves average GLUE validation score on par with full fine-tuning while fine-tuning approximately $100$ times fewer parameters.

\subsection{Very Small PETs}

Here we examine our learned PET architectures with parameter budget less than the total number of bias parameters in the base PLM. For \verb|roberta-large|, this is about 273k.

We use our method to learn structured and unstructured bias-tuning architectures. We compare our method with WARP \cite{hambardzumyan2021warp} using parameter budget 25k in Table \ref{table:results_table}, and report results for our method for other parameter budgets in the appendix. Our learned structured and unstructured bias-tuning architectures are labeled \verb|S-BitFit| and \verb|U-BitFit|, respectively. In our \verb|S-BitFit| and \verb|U-BitFit| experiments, we prune from a PET architecture with 273k parameters that fine-tuens all bias parameters, the same as BitFit. We observe that the unstructured bias-tuning architecture achieves significantly higher validation performance than the structured bias-tuning architecture with the same parameter budget. We conclude that the subset of bias parameters that are "good" to fine-tune are not concentrated in a few modules, but rather are distributed throughout the network. Our learned unstructured bias-tuning architecture with $<50$k parameters fine-tunes only $18\%$ of all bias parameters while achieving validation GLUE score only slightly less than fine-tuning all bias parameters ($86.5$ versus $86.7$). We conclude that a vast majority of bias parameters do not need to be fine-tuned to achieve performance comparable to fine-tuning all bias parameters. With a parameter budget of 25k, unstructured bias tuning achieves similar performance compared to WARP, beating or tying WARP on a majority of GLUE tasks but achieving slightly worse average performance. We conclude that both methods are about equally effective.

\subsection{Interpreting Learned Architectures}

Here we examine the architectures learned by our algorithm and consider what they say about which parts of the network are most parameter-efficient to fine-tune. Each illustration discussed in this section averages the architectures learned by our method over all GLUE tasks and five random initializations per task. Figure \ref{fig:sbt_architecture} illustrates the architecture learned by our method for structured bias-tuning with parameter budget 50k. We observe a clear preference by our algorithm for fine-tuning the biases of the \verb|intermediate.dense| modules in the middle of the network. Figure \ref{fig:ubt_architecture} illustrates the architecture learned by our method for unstructured bias tuning with parameter budget 50k. We observe a weak preference for fine-tuning the bias parameters of modules in the middle of the network, but not for any particular module type within each transformer block. We conclude that the biases that are most parameter-efficient to fine-tune are in the middle layers of the network.

\section{Conclusion}
In this paper, we considered the question \textit{which parts of the network are most efficient to fine-tune, and what is the most parameter-efficient way to fine-tune them}? To answer that question, we developed a NAS algorithm based on structured and unstructured pruning. We presented experimental results on RoBERTa Large demonstrating the effectiveness of our algorithm, achieving GLUE validation performance similar to WARP at 25k parameters (9\% of all biases), similar to BitFit at 50k parameters (18\% of all biases), and similar to full fine-tuning at 3.4M parameters (10\% of all parameters). From our learned architectures we observed that the bias parameters in the middle layers of the network are most efficient to fine-tune. We conclude that it is important to consider \textit{where} to fine-tune as well as \textit{how}.


\section*{Limitations}
Differences in experimental setup may make it difficult to accurately and fairly compare published results. For example, to prevent data leakage, we report validation performance at the end of training and do not perform early stopping. This is in contrast to most other papers which report peak validation performance. Results reported for other methods are reproduced in the same learning environment as our method unless explicitly stated otherwise. This takes into account recent work demonstrating problems with fairly and accurately evaluating PET methods that use early stopping improperly \cite{chen2022revisiting}.

Although many pruning criteria exist in the literature, in this paper we only consider one pruning criterion. Although not presented in this paper, experiments we conducted with various formulations of magnitude pruning did not produce better results.

Although prompt tuning is a popular PET method, we do not perform NAS for prompt tuning to determine the most efficient positions for inserting prompt tokens into the input. Pruning may or may not prove to be a successful strategy for this problem.

Other NAS strategies exist in the literature besides pruning, such as evolutionary, reinforcement learning, and DARTS \cite{liu2018darts}. However, our pruning method seems to give a good trade-off between validation performance and computational expense.

\section*{Ethics Statement}
Powerful language models can be used for unethical purposes, such as generating offensive or deceptive content. Although researchers today are making a greater effort to establish protections against the unethical use of their models, bad actors may still find ways to circumvent those protections. One avenue for attack could involve fine-tuning a PLM on a nefarious dataset to produce unethical content. In this paper, we showed that a PLM can be successfully fine-tuned on a downstream task by fine-tuning a small number of parameters, or adding a low-rank update to a few select parameter matrices. Thus researchers should consider the risk posed by unethical parameter-efficient fine-tuning before publishing a fine-tuneable version of their model.


\bibliography{main}
\bibliographystyle{acl_natbib}

\appendix
\label{sec:appendix}

\begin{table*}[hbt!]
\centering
\begin{tabular}{|c|c|c|c|c|c|c|c|c|c|c|c|}
\hline
Method & \#params & MNLI & SST-2 & MRPC & CoLA & QNLI & QQP & RTE & STS-B & Avg. \\
\hline
WARP$^\dagger$ & 11k & $87.6$ & $93.0$ & $83.8$ & $72.9$ & $95.4$ & $85.6$ & $57.4$ & $81.0$ & $82.1$ \\
WARP$^\dagger$ & 25k & $88.2$ & $96.0$ & $90.8$ & $60.6$ & $93.5$ & $84.5$ & $75.8$ & $88.6$ & $84.8$ \\
\hline
S-BitFit & 10k & $70.1$ & $92.1$ & $70.6$ & $0.0$ & $73.1$ & $73.3$ & $52.7$ & $22.2$ & $56.8$
 \\
S-BitFit & 25k & $84.1$ & $94.2$ & $70.6$ & $40.2$ & $88.9$ & $83.8$ & $56.0$ & $76.8$ & $74.3$ \\
S-BitFit & 50k & $87.1$ & $94.3$ & $72.1$ & $51.5$ & $91.4$ & $86.2$ & $59.6$ & $86.9$ & $78.6$ \\
S-BitFit & 100k & $88.2$ & $95.0$ & $87.7$ & $58.8$ & $92.4$ & $87.4$ & $78.7$ & $90.4$ & $84.8$ \\
S-BitFit & 200k & $89.1$ & $95.6$ & $88.2$ & $63.1$ & $93.8$ & $87.9$ & $81.9$ & $91.4$ & $86.4$ \\
\hline
U-BitFit & 10k & $87.4$ & $95.1$ & $71.1$ & $58.8$ & $92.2$ & $86.3$ & $59.6$ & $88.3$ & $79.8$
 \\
 U-BitFit & 25k & $88.8$ & $95.5$ & $85.3$ & $62.1$ & $93.5$ & $87.7$ & $74.0$ & $90.3$ & $84.6$ \\
U-BitFit & 50k & $89.1$ & $95.8$ & $88.5$ & $64.8$ & $93.8$ & $88.0$ & $80.9$ & $91.1$ & $86.5$ \\
U-BitFit & 100k & $89.3$ & $95.8$ & $88.5$ & $63.6$ & $93.9$ & $87.7$ & $81.9$ & $91.3$ & $86.5$ \\
U-BitFit & 200k & $89.4$ & $95.6$ & $88.5$ & $64.8$ & $93.9$ & $86.5$ & $81.9$ & $91.4$ & $86.5$
 \\
\hline
\end{tabular}
\caption{GLUE development set score for structured (S-BitFit) and unstructured (U-BitFit) bias-tuning architectures learned by our method for different parameter budgets. The results for WARP$^\dagger$ are reported from \citet{hambardzumyan2021warp}.}
\label{table:bias_tuning_parameter_budgets}
\end{table*}

\section{Experiment Setup}

In all experiments we use the Adam optimizer \cite{kingma2014adam} and a linear learning rate scheduler with $6\%$ warm-up steps. We observe that training with a higher peak learning rate works best when fine-tuning a small number of parameters. We use different peak learning rates for different experiments depending on the maximum number of parameters being fine-tuned, ranging from $10^{-5}$ for full fine-tuning to $3 \times 10^{-4}$ for training our smallest PETs. We also train for a different number of epochs for each GLUE tasks. We train for $20$ epochs on MRPC, RTE, CoLA, and STSB; $5$ epochs on SST-2 and QNLI; and $2$ epochs for MNLI and QQP. We observe that extending the number of training epochs beyond these limits does not substantially affect validation performance. In all experiments, we use batch size $16$ and maximum sequence length $128$.

\section{Additional Experimental Results}
We report results for our learned structured and unstructured bias-tuning architecture with parameter budgets 10k, 25k, 50k, 100k, and 200k in Table \ref{table:bias_tuning_parameter_budgets}. We observe that unstructured bias-tuning holds an advantage over structured bias-tuning across all parameter budgets. We also observe that the performance of unstructured bias-tuning begins to fall off after decreasing the parameter budget below 50k. WARP with a parameter budget of 11k significantly outperforms our \verb|U-BitFit| method with a parameter budget of 10k on the MRPC and COLA tasks. This difference might be explained by the difference in experimental setup (e.g., \citet{hambardzumyan2021warp} reports peak validation score whereas we report end-of-training validation score), or the small difference in parameter budget. We believe that our method can be improved in the very small parameter budget regime using iterative, rather than one-shot, pruning.

\end{document}